\pdfoutput=1

\documentclass[11pt]{article}

\usepackage[]{acl}

\usepackage{times}
\usepackage{latexsym}

\usepackage[T1]{fontenc}

\usepackage[utf8]{inputenc}

\usepackage{microtype}

\usepackage{inconsolata}

\usepackage{adjustbox}
\usepackage{amsmath}
\usepackage[amssymb]{SIunits}
\usepackage{amssymb}

\usepackage{algorithm}

\usepackage{supertabular}
\usepackage{array}
\usepackage{psfrag}
\usepackage{hhline}
\usepackage{type1cm}
\usepackage[normalem]{ulem}
\usepackage[noend]{algpseudocode}
\usepackage{pgfplots}
\pgfplotsset{compat=1.14}

\usepackage{ctable}
\usepackage{tikz}
\usetikzlibrary{positioning,arrows,trees,shapes,fit}
\usepackage{tikz-cd}
\usetikzlibrary{backgrounds}

\tikzset{
    >=stealth',
    punkt/.style={
           rectangle,
           rounded corners,
           draw=black, very thick,
           text width=6.5em,
           minimum height=2em,
           text centered},
    pil/.style={
           ->,
           thick,
           shorten <=2pt,
           shorten >=2pt,}
}

\usepackage{multicol}
\usepackage{multirow}
\usepackage{wrapfig}
\usepackage{enumerate}
\usepackage{enumitem}
\usepackage{subcaption}
\usepackage{mathtools}
\usepackage{lipsum}
\usepackage{paracol}
\usepackage{pifont}
\usepackage{verbatim}
\usepackage[utf8]{inputenc}
\usepackage[T2A, T5, T1]{fontenc}

\usepackage{CJKutf8}
\usepackage{color,soul}               

\usepackage[main=english,russian,vietnamese]{babel}

\usepackage{pgfplotstable} 

\usepackage{graphicx}

\definecolor{LRed}{rgb}{1,.9,.9}
\definecolor{LGreen}{rgb}{.9,1,0.9} 
\definecolor{LBlue}{rgb}{.9,.9,1}
\definecolor{LYellow}{rgb}{1,1,0.9}

\definecolor{lightblue}{rgb}{0.68, 0.85, 0.9}
\definecolor{lavender}{rgb}{0.9, 0.9, 0.98}
\definecolor{lightyellow}{rgb}{1.0, 1.0, 0.88}
\definecolor{magicmint}{rgb}{0.67, 0.94, 0.82}
\definecolor{palepink}{rgb}{0.98, 0.85, 0.87}
\definecolor{bubbles}{rgb}{0.91, 1.0, 1.0}

\usepackage{float}

\usepackage{enumerate}

\definecolor{darkblue}{rgb}{0.0, 0.0, 0.5}
\definecolor{airforceblue}{rgb}{0.0, 0.0, 0.8}
\definecolor{crimson}{rgb}{0.8, 0.0, 0.0}

\newcommand{\ie}{\emph{i.e.,}}
\newcommand{\eg}{\emph{e.g.,}}

\usepackage{amsmath,amsfonts,bm}

\def\eqref#1{equation~\ref{#1}}

\def\1{\bm{1}}


\DeclareMathAlphabet{\mathsfit}{\encodingdefault}{\sfdefault}{m}{sl}
\SetMathAlphabet{\mathsfit}{bold}{\encodingdefault}{\sfdefault}{bx}{n}

\pgfplotstableread{
id    lang  percent 
1       tu	0.00002
2       ki	0.00004
3       bm	0.00004
4       ak	0.00007
5       tg	0.00007
6       ss	0.00007
7       ny	 0.0001
8       tn	 0.0002
9       ln	 0.0002
10      ns	 0.0002
11      fo	 0.0002
12      rd	 0.0003
13      wo	 0.0004
14      lu	 0.0004
15      sn	  0.001
16      zu	  0.001
17      ig	  0.001
18      xh	  0.001
19      kw	  0.003
20      yo	  0.006
21      sw	   0.02
22      as	   0.01
23      or	   0.04
24      gu	   0.04
25      mr	   0.05
26      pa	   0.05
27      kn	   0.06
28      ne	   0.07
29      te	   0.09
30      ml	    0.1
31      ur	    0.1
32      ta	    0.2
33      bn	    0.5
34      hi	    0.7
}\langpercent

\pgfplotstableread{
id    lang  percent 
1       tu	0.00002
2       ki	0.00004
3       bm	0.00004
4       ak	0.00007
5       tg	0.00007
6       ss	0.00007
7       ny	 0.0001
8       tn	 0.0002
9       ln	 0.0002
10      ns	 0.0002
11      fo	 0.0002
12      rd	 0.0003
13      wo	 0.0004
14      lu	 0.0004
15      sn	  0.001
16      zu	  0.001
17      ig	  0.001
18      xh	  0.001
19      kw	  0.003
20      yo	  0.006
21      sw	   0.02
}\langpercentafrican

\pgfplotstableread{
id    lang  percent 
22      as	   0.01
23      or	   0.04
24      gu	   0.04
25      mr	   0.05
26      pa	   0.05
27      kn	   0.06
28      ne	   0.07
29      te	   0.09
30      ml	    0.1
31      ur	    0.1
32      ta	    0.2
33      bn	    0.5
34      hi	    0.7
}\langpercentindic

\pgfplotstableread{
1   53.0  36.027  8.219   51.0   99.726  1
2   56.0  21.918  4.110   53.0   12.329  1
3   56.0  16.438  6.849   55.0   39.726  1
4   64.0  2.740   2.740   56.0   28.767  1
5   55.0  1.370   6.849   52.0   1.3700  2
6   58.0  2.740   16.43   52.0   57.534  2
7   58.0  0.000   0.000   55.0   32.877  2
8   63.0  6.849   5.479   56.0   15.068  2
9   56.0  12.329  6.849   50.0   20.548  3
10  54.0  4.110   8.219   52.0   35.616  3
11  58.0  12.329  6.849   54.0   20.548  3
12  64.0  4.110   8.219   57.0   35.616  3
}\dataA

\pgfplotstableread{
id    lang  percent  sup_toen  sup_fromen  our_toen  our_fromen
1       as	    0.01	  46.21	      27.13	     46.78		  26.57
2       or	    0.04	  49.99	      32.00	     50.86		  31.24
3       gu	    0.04	  51.63	      38.23	     50.09		  37.63
4       mr	    0.05	  33.08	      25.73	     33.70		  24.97
5       pa	    0.05	  54.04	      31.99	     53.35		  32.27
6       kn	    0.06	  47.82	      34.17	     47.94		  33.48
7       ne	    0.07	  47.07	      35.91	     48.26		  35.76
8       te	    0.09	  44.44	      36.70	     44.85		  37.26
9       ml	     0.1	  38.67	      25.56	     35.47		  25.39
10      ur	     0.1	  50.20	      37.00	     52.36		  37.45
11      ta	     0.2	  43.93	      38.63	     46.47		  38.80
12      bn	     0.5	  52.82	      42.56	     53.25		  42.78
13      hi	     0.7	  55.18	      44.94	     55.71		  45.36
}\dataindic

\pgfplotstableread{
id    lang  percent  percent2  sup_toen  sup_fromen  our_toen  our_fromen
1       tu	0.00002     0.002	    26.43		   9.89	      26.39	    10.91
2       ki	0.00004     0.004	    25.26		  11.59	      20.51	    9.12
3       bm	0.00004     0.004	    20.69		   6.35	      21.73	    8.45
4       ak	0.00007     0.007	    25.97		  12.67	      25.15	    12.23
5       tg	0.00007     0.007	    24.69		  11.86	      25.67	    9.75
6       ss	0.00007     0.007	    24.53		  13.04	      21.61	    11.28
7       ny	 0.0001      0.01	    27.74		  12.85	      28.91	    12.96
8       tn	 0.0002      0.02	    24.69		  14.48	      26.38	    14.18
9       ln	 0.0002      0.02	    28.73		  17.92	      28.45	    16.95
10      ns	 0.0002      0.02	    26.42		  12.84	      27.23	    12.28
11      fo	 0.0002      0.02	    19.73		   5.12	      19.45	    4.84
12      rd	 0.0003      0.03	    35.16		  16.36	      34.54	    15.22
13      wo	 0.0004      0.04	    27.27		  14.14	      28.63	    13.61
14      lu	 0.0004      0.04	    26.30		  13.68	      27.35	    12.33
15      sn	  0.001       0.1	    27.44		  12.52	      27.68	    12.10
16      zu	  0.001       0.1	    23.45		   9.56	      21.98	    9.51
17      ig	  0.001       0.1	    32.63		  17.61	      33.87	    17.46
18      xh	  0.001       0.1	    25.89		  12.53	      26.56	    13.19
19      kw	  0.003       0.3	    40.10		  20.52	      40.14	    20.71
20      yo	  0.006       0.6	    31.78		  20.21	      34.46	    21.93
21      sw	   0.02         2	    56.51		  47.68	      56.42	    47.06
}\dataafrican

\pgfplotstableread{
id    lang  percent   sup_toen  sup_fromen  our_toen  our_fromen
1       tu	0.00002 	  26.43		 9.89		26.39		10.91
2       ki	0.00004 	  25.26		11.59		20.51		9.12
3       bm	0.00004 	  20.69		 6.35		21.73		8.45
4       ak	0.00007 	  25.97		12.67		25.15		12.23
5       tg	0.00007 	  24.69		11.86		25.67		9.75
6       ss	0.00007 	  24.53		13.04		21.61		11.28
7       ny	 0.0001 	  27.74		12.85		28.91		12.96
8       tn	 0.0002 	  24.69		14.48		26.38		14.18
9       ln	 0.0002 	  28.73		17.92		28.45		16.95
10      ns	 0.0002 	  26.42		12.84		27.23		12.28
11      fo	 0.0002 	  19.73		 5.12		19.45		4.84
12      rd	 0.0003 	  35.16		16.36		34.54		15.22
13      wo	 0.0004 	  27.27		14.14		28.63		13.61
14      lu	 0.0004 	  26.30		13.68		27.35		12.33
15      sn	  0.001 	  27.44		12.52		27.68		12.10
16      zu	  0.001 	  23.45		 9.56		21.98		9.51
17      ig	  0.001 	  32.63		17.61		33.87		17.46
18      xh	  0.001 	  25.89		12.53		26.56		13.19
19      kw	  0.003 	  40.10		20.52		40.14		20.71
20      yo	  0.006 	  31.78		20.21		34.46		21.93
21      sw	   0.02 	  56.51		47.68		56.42		47.06
22      as	   0.01	    46.21		 27.13	 46.78	26.57
23      or	   0.04	    49.99		 32.00	 50.86	31.24
24      gu	   0.04	    51.63		 38.23	 50.09	37.63
25      mr	   0.05	    33.08		 25.73	 33.70	24.97
26      pa	   0.05	    54.04		 31.99	 53.35	32.27
27      kn	   0.06	    47.82		 34.17	 47.94	33.48
28      ne	   0.07	    47.07		 35.91	 48.26	35.76
29      te	   0.09	    44.44		 36.70	 44.85	37.26
30      ml	    0.1	    38.67		 25.56	 35.47	25.39
31      ur	    0.1	    50.20		 37.00	 52.36	37.45
32      ta	    0.2	    43.93		 38.63	 46.47	38.80
33      bn	    0.5	    52.82		 42.56	 53.25	42.78
34      hi	    0.7	    55.18		 44.94	 55.71	45.36
}\dataafricanindic

\pgfplotstableread{
id    lang  percent   sup_toen  sup_fromen  our_toen  our_fromen	supfen_m_ourfen   bloom_tok_ratio     openai_tok_ratio    openai_sup_toen     openai_sup_fromen   openai_m_blo_toen	openai_m_blo_fromen
1       tu	0.00002 	26.43		  9.89	 26.39	10.91		-1.02                           2.210               2.802               29.62          	      14.75           3.19	        4.86
2       ki	0.00004 	25.26		 11.59	 20.51	9.12		 2.47                           2.527               3.504               25.10          	      13.45           -0.16	        1.86
3       bm	0.00004 	20.69		  6.35	 21.73	8.45		 -2.1                           1.917               2.687               25.10          	      11.69           4.41	        5.34
4       ak	0.00007 	25.97		 12.67	 25.15	12.23		 0.44                           2.067               2.803               26.11          	      24.56           0.14	        11.89
5       tg	0.00007 	24.69		 11.86	 25.67	9.75		 2.11                           2.026               2.471               33.95          	      22.88           9.26	        11.02
6       ss	0.00007 	24.53		 13.04	 21.61	11.28		 1.76                           1.984               2.365               25.10          	      12.93           0.57	        -0.11
7       ny	 0.0001 	27.74		 12.85	 28.91	12.96		-0.11                           1.810               2.295               30.29          	      20.50           2.55	        7.65
8       tn	 0.0002 	24.69		 14.48	 26.38	14.18		  0.3                           2.063               2.434               26.12          	      20.13           1.43	        5.65
9       ln	 0.0002 	28.73		 17.92	 28.45	16.95		 0.97                           1.662               2.042               30.42          	      23.92           1.69	        6
10      ns	 0.0002 	26.42		 12.84	 27.23	12.28		 0.56                           1.960               2.343               28.56          	      21.70           2.14	        8.86
11      fo	 0.0002 	19.73		  5.12	 19.45	4.84		 0.28                           2.220               4.100               24.61          	      8.260           4.88	        3.14
12      rd	 0.0003 	35.16		 16.36	 34.54	15.22		 1.14                           1.653               2.337               30.15          	      17.12           -5.01	        0.76
13      wo	 0.0004 	27.27		 14.14	 28.63	13.61		 0.53                           1.690               2.154               28.84          	      19.77           1.57	        5.63
14      lu	 0.0004 	26.30		 13.68	 27.35	12.33		 1.35                           1.690               2.191               28.84          	      19.77           2.54	        6.09
15      sn	  0.001 	27.44		 12.52	 27.68	12.10		 0.42                           1.826               2.322               35.82          	      24.23           8.38	        11.71
16      zu	  0.001 	23.45		  9.56	 21.98	9.51		 0.05                           1.761               2.416               27.12          	      18.40           3.67	        8.84
17      ig	  0.001 	32.63		 17.61	 33.87	17.46		 0.15                           1.703               3.397               29.10          	      19.12           -3.53	        1.51
18      xh	  0.001 	25.89		 12.53	 26.56	13.19		-0.66                           1.691               2.298               25.00          	      14.73           -0.89	        2.2
19      kw	  0.003 	40.10		 20.52	 40.14	20.71		-0.19                           1.602               2.398               30.11          	      15.77           -9.99	        -4.75
20      yo	  0.006 	31.78		 20.21	 34.46	21.93		-1.72                           1.671               3.899               29.11          	      16.71           -2.67	        -3.5
21      sw	   0.02 	56.51		 47.68	 56.42	47.06		 0.62                           1.251               2.147               51.08          	      43.17           -5.43	        -4.51
22      as	   0.01	    46.21		 27.13	 46.78	26.57		 0.56                           1.439               9.878               33.96          	      19.06           -12.25	    -8.07
23      or	   0.04	    49.99		 32.00	 50.86	31.24		 0.76                           1.378               13.47               37.81          	      16.67           -12.18	    -15.33
24      gu	   0.04	    51.63		 38.23	 50.09	37.63		  0.6                           1.369               12.30               40.79          	      23.22           -10.84	    -15.01
25      mr	   0.05	    33.08		 25.73	 33.70	24.97		 0.76                           1.230               7.897               26.12          	      28.53           -6.96	        2.8
26      pa	   0.05	    54.04		 31.99	 53.35	32.27		-0.28                           1.453               7.932               40.09          	      26.38           -13.95	    -5.61
27      kn	   0.06	    47.82		 34.17	 47.94	33.48		 0.69                           1.345               13.75               36.40          	      20.01           -11.42	    -14.16
28      ne	   0.07	    47.07		 35.91	 48.26	35.76		 0.15                           1.196               7.679               36.54          	      25.34           -10.53	    -10.57
29      te	   0.09	    44.44		 36.70	 44.85	37.26		-0.56                           1.361               13.17               27.88          	      24.66           -16.56	    -12.04
30      ml	    0.1	    38.67		 25.56	 35.47	25.39		 0.17                           1.393               15.32               26.84          	      15.11           -11.83	    -10.45
31      ur	    0.1	    50.20		 37.00	 52.36	37.45		-0.45                           1.386               6.348               45.61          	      34.56           -4.59	        -2.44
32      ta	    0.2	    43.93		 38.63	 46.47	38.80		-0.17                           1.298               15.64               29.10          	      18.84           -14.83	    -19.79
33      bn	    0.5	    52.82		 42.56	 53.25	42.78		-0.22                           1.190               9.723               47.11          	      29.67           -5.71	        -12.89
34      hi	    0.7	    55.18		 44.94	 55.71	45.36		-0.42                           1.308               7.520               53.61          	      39.62           -1.57	        -5.32
}\dataafricanindic

\pgfplotstableread{

id  lorar		params_n    indic2en    en2indic  unsup_indic2en unsup_en2indic x2en_diff   en2x_diff
1		8	      $3e^7$	41.19		23.98		    39.88		24.41       1.31	-0.43
2		16	      $5e^7$	41.25		26.01		    39.88		24.41       1.37	1.6
3		32	      $1e^8$	41.53		27.31		    39.88		24.41       1.65	2.9
4		64	      $2e^8$	41.59		28.69		    39.88		24.41       1.71	4.28
5		128	      $5e^8$	41.71		29.69		    39.88		24.41       1.83	5.28
6		256	      $1e^9$	41.89		31.51		    39.88		24.41       2.01	7.1
7		512	      $2e^9$	42.19		32.72		    39.88		24.41       2.31	8.31
}\datafinetuneindic

\usepackage{hyperref}
\usepackage{cleveref}
\usepackage{url}

\usepackage{microtype}

\def\logo{\makebox[0pt][l]{\hspace{0pt}\raisebox{-1ex}{\includegraphics[height=32pt]{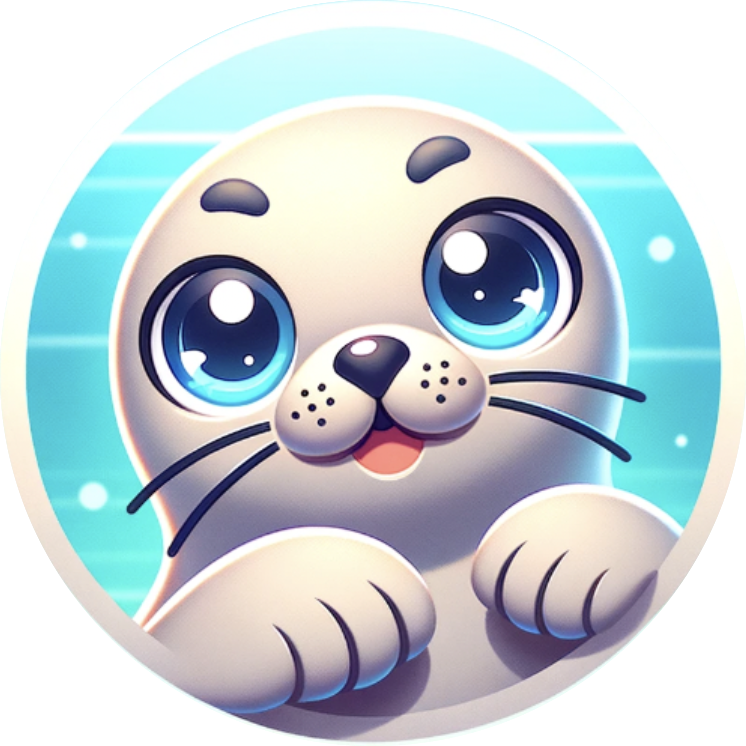}}}}
\title{\logo \ \ \ \ \ \ \ \ \ SeaLLMs - Large Language Models for Southeast Asia}

\author{%
    Xuan-Phi Nguyen\thanks{\ \ $^{\ddagger}$ Equal contributions.}, Wenxuan Zhang$^*$, Xin Li$^*$, Mahani Aljunied$^*$, Zhiqiang Hu, \\
    \textbf{Chenhui Shen$^{\ddagger}$, Yew Ken Chia$^{\ddagger}$, Xingxuan Li, Jianyu Wang, Qingyu Tan, Liying Cheng,} \\
    \textbf{Guanzheng Chen, Yue Deng, Sen Yang, Chaoqun Liu, Hang Zhang, Lidong Bing\thanks{\ \  Corresponding author: \href{mailto:l.bing@alibaba-inc.com}{l.bing@alibaba-inc.com}}}\\\\
  DAMO Academy, Alibaba Group \\
  Video: \url{https://youtu.be/s0mBrHYD_H4}\\
  Website: \url{https://damo-nlp-sg.github.io/SeaLLMs}
}

\begin{document}
\maketitle
\begin{abstract}

Despite the remarkable achievements of large language models (LLMs) in various tasks, there remains a linguistic bias that favors high-resource languages, such as English, often at the expense of low-resource and regional languages. To address this imbalance, we introduce SeaLLMs, an innovative series of language models that specifically focuses on Southeast Asian (SEA) languages. SeaLLMs are built upon popular English-centric models through continued pre-training with an extended vocabulary, specialized instruction and alignment tuning to better capture the intricacies of regional languages. This allows them to respect and reflect local cultural norms, customs, stylistic preferences, and legal considerations. Our comprehensive evaluation demonstrates that SeaLLM models exhibit superior performance across a wide spectrum of linguistic tasks and assistant-style instruction-following capabilities relative to comparable open-source models. Moreover, they outperform ChatGPT-3.5 in non-Latin languages, such as Thai, Khmer, Lao, and Burmese, by large margins while remaining lightweight and cost-effective to operate. 
\end{abstract}


\section{Introduction}\label{sec:intro}




\begin{figure*}[t] 
    \centering
    \includegraphics[width=0.9\textwidth]{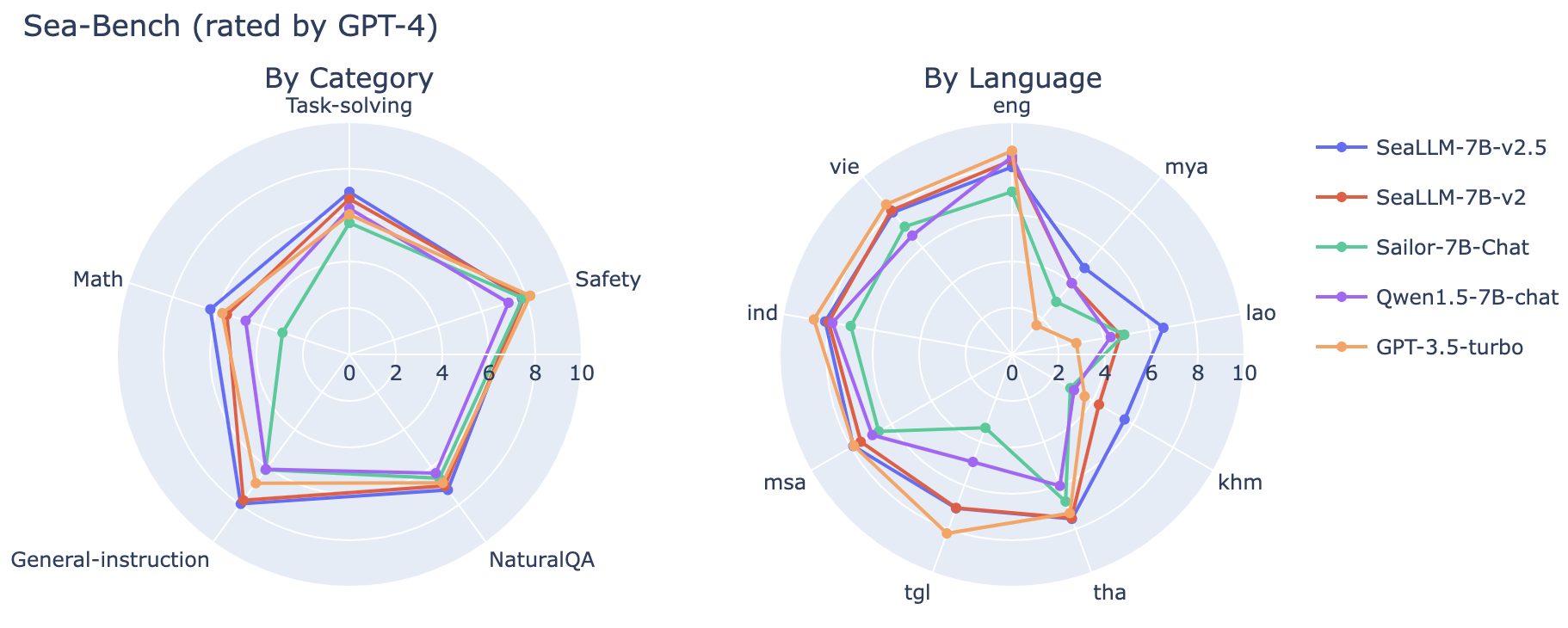} 
    \caption{Sea-bench (\Cref{sec:eval:peer_com}) scores as evaluated by GPT-4 \citep{mt_bench_zheng2023judging} for different models. 
    Each radar chart compares scores as averaged across 5 categories (left) and 9 languages (right).
        Detailed breakdown by each category and language is given in \Cref{fig:sea_bench:seallm_vs_chatgpt_breakdown} in the Appendix.
    } 
    \label{fig:sea_bench:seallm_vs_chatgpt}
\end{figure*}

The advent of large language models (LLMs) has radically transformed the field of natural language processing, demonstrating remarkable abilities in text generation, comprehension, and decision-making tasks \cite{gpt3_brown2020language,chatgpt,gpt4,llama_touvron2023,llama2touvron2023llama,lambda_google_thoppilan2022lamda,mistral7b_jiang2023mistral,polylm_wei2023polylm,qwen_bai2023}. While the proficiencies of these models are extraordinary, the majority of existing LLMs embody a linguistic hierarchy overwhelmingly dominated by English \cite{mega, multilingual-eval, m3exam}. This dominance undermines the multilingual capability of such models, with particularly prejudicial outcomes for lower-resource and regional languages, where data scarcity and tokenization challenges lead to disproportionately poor model performance. This linguistic disparity not only impedes access to state-of-the-art AI technologies for non-English-speaking populations but also risks cultural homogenization and the loss of linguistic diversity. While hyper-polyglot models exist \citep{bloom_scao2022bloom,bloomz_muennighoff2022crosslingual,polylm_wei2023polylm}, they may pay a high cost for high-resource language performance while lacking in multilingual instruction-following abilities.

Recognizing the urgent need to democratize AI and empower linguistically diverse regions, we introduce SeaLLMs\footnote{\href{https://github.com/DAMO-NLP-SG/SeaLLMs}{https://github.com/DAMO-NLP-SG/SeaLLMs}}, a suite of specialized language models optimized for Southeast Asian languages\footnote{English (Eng), Chinese (Zho), Indonesian (Ind), Vietnamese (Vie), Thai (Tha), Khmer (Khm), Lao, Malay (Msa), Burmese (Mya) and Tagalog (Tgl)}. These languages, while rich and diverse, often lack the extensive dataset support available for more widely spoken languages, resulting in a stark performance gap in existing LLM applications. 

As a long-term continuous effort, as of this writing, SeaLLMs come in three versions (v1, v2, v2.5). SeaLLM-13B-v1, which was pre-trained from Llama-2-13B, eclipses the performance of most available open-source LLMs in a comprehensive array of tasks including world knowledge assessments, language comprehension, and generative capabilities in SEA languages. For English and alike, SeaLLMs do not only preserve, but also demonstrate enhanced performance in tasks that were part of the original Llama training set. When evaluated on multilingual instruction-following tasks with GPT-4 as a judge \citep{mt_bench_zheng2023judging}, SeaLLM-13B-v1 outperforms ChatGPT-3.5 by large margins in less-represented languages such as Khmer, Lao or Burmese. Meanwhile, SeaLLM-7B-v2, which was pre-trained from Mistral-7B \citep{mistral7b_jiang2023mistral}, demonstrates better performances in math and commonsense reasoning than comparable baselines, surpassing ChatGPT-3.5 in reasoning for common SEA languages, while being much smaller in sizes. Later, SeaLLM-7B-v2.5, which was further pre-trained from Gemma-7B \citep{gemma_team2024}, shows significant improvements in SEA languages over SeaLLM-7B-v2.

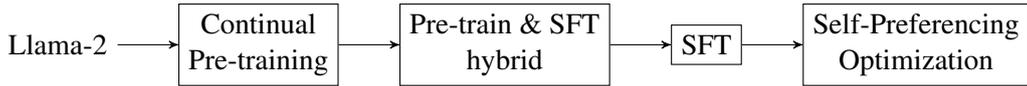
\begin{figure*}[h]
  \centering
    \begin{tikzpicture}[node distance=0.8cm, auto]
      \node (llama2) {Llama-2};
      \node (pretrain) [right=of llama2, rectangle, draw, align=center] {Continual\\Pre-training};
      \node (pretrainHybrid) [right=of pretrain, rectangle, draw, align=center] {Pre-train \& SFT\\hybrid};
      \node (sft) [right=of pretrainHybrid, rectangle, draw] {SFT};
      \node (dpo) [right=of sft, rectangle, draw, align=center] {Self-Preferencing\\Optimization};
      \draw[->] (llama2) -- (pretrain);
      \draw[->] (pretrain) -- (pretrainHybrid);
      \draw[->] (pretrainHybrid) -- (sft);
      \draw[->] (sft) -- (dpo);
    \end{tikzpicture}
  \caption{Complete Training Process of SeaLLMs. It begins with continual pre-training Llama-2 with more data of regional languages. Then the models undergo specialized fine-tuning process with multilingual SFT data, before finally being tuned with self-preferencing alignment.}
  \label{fig:seallm_train_process}
\end{figure*}

\Cref{fig:seallm_train_process} illustrates the four-stage training process of SeaLLMs. In the first stage, detailed in \Cref{sec:pretrain:process}, we conduct continuous pre-training from the foundational models \citep{llama2touvron2023llama,mistral7b_jiang2023mistral} with an extended vocabulary tailored for SEA languages. Next, we fine-tune the model in a novel hybrid paradigm with a mixture of multilingual pre-training data and English-dominant instruction fine-tuning data (\Cref{sec:sft:pretrain_sft}). The following stage subsequently fine-tunes the model on a balanced and custom-built multilingual SFT dataset. Finally, we conduct self-preferencing alignment optimization using the SeaLLM model itself, without relying on human annotators or more powerful LLMs \citep{gpt4}.

\section{Pre-training}\label{sec:pretraining}

\subsection{Pre-training Data}\label{sec:pretrain:data}


The pre-training data comprises a heterogeneous assortment of documents sourced from several publicly accessible repositories~\citep{suarez2019asynchronous,mc4_2019t5,together2023redpajama,wikidump}. Specifically, during the creation of the pre-training data, we include web-based corpora such as Common Crawl \citep{ccnet_wenzek-etal-2020-ccnet}, journalistic content such as CC-News, text corpora with expertly-curated knowledge such as Wikipedia \citep{wikidump}, and some scholarly publications. After collecting the data, we employ a language identifier \citep{fasttext} to retain the documents for the major languages in Southeast Asia, namely Thai, Vietnamese, Indonesian, Chinese, Khmer, Lao, Malay, Burmese, and Tagalog, and discard the remaining ones. Subsequent stages of data refinement involve the multiple modules dedicated to data cleansing and content filtration.
We blend such data with the highest quality English data from RedPajama subset \citep{together2023redpajama} in more balanced ratios, as we found that such English data are useful to preserve the original learnt knowledge.

\subsection{Vocabulary Expansion}\label{sec:pretrain:vocab_extend}

\begin{table}[t]
\centering
\resizebox{\columnwidth}{!}{%
\begin{tabular}{lccc}
\toprule
Language & ChatGPT's & Llama's & SeaLLM's \\
\midrule
\quad Vie &  4.41 &  3.46 & 1.48 \\
\quad Zho &  2.80 &  2.36 & 1.40 \\
\quad Tha &  9.09 &  5.10 & 1.87 \\
\quad Ind &  2.00 &  2.09 & 1.36 \\
\quad Khm & 15.56 & 12.14 & 2.67 \\
\quad Lao & 13.29 & 13.50 & 2.07 \\
\quad Msa &  2.07 &  2.16 & 1.50 \\
\quad Mya & 17.11 &  9.85 & 1.93 \\
\quad Tgl &  2.28 &  2.22 & 1.91 \\
\midrule
\quad Eng &  {1.00 (baseline)} &  1.19 & 1.19 \\
\bottomrule
\end{tabular}
}
\caption{Averaged compression ratios between the tokenized length of texts of each language produced by different tokenizers versus the baseline tokenized length of same-meaning English equivalents produced by ChatGPT tokenizer (\ie\ it costs 15.6x more tokens to encode Khmer than English with ChatGPT tokenizer). SeaLLM's ratios are applicable only for v1 and v2.}
\label{tab:language_ratios}
\end{table}


\Cref{tab:language_ratios} describes how expensive it is to process an under-represented non-Latin language. For example, encoding a single sentence in Thai requires 4.3 times more tokens than its English equivalent.
The reason for this is that most English language models employ a BPE tokenizer \citep{bpe_sennrich2015neural} that inefficiently segments texts from non-Latin scripts into disproportionately lengthy byte sequences, which inadequately represent the underlying semantic content, resulting in diminished model performance \citep{democratize_llm_ldp_nguyen2023}. 
To that end, we propose a novel vocabulary expansion technique, as formally described in \Cref{algo:vocab_extend} in the Appendix. This technique involves recursively merging whole-word and sub-word token pieces of a new language from a highly multilingual target tokenizer (\ie\ the NLLB tokenizer \citep{nllb_costa2022no_flores200}), to the existing LLM tokenizer. This new set of retrieved tokens are then pruned to remove rarely appearing and low-quality tokens before being added to the final SeaLLM tokenizer.

\Cref{tab:language_ratios} demonstrates the efficiency of the new vocabulary. The compression ratio for Thai text has markedly improved from 4.29 to 1.57, signifying a 2.7-fold increase in the length of Thai text that can be encoded within the same context constraints. At the same time, the compression of English text has experienced a negligible reduction of 0.3\%, thus maintaining its tokenization effectiveness.

We applied our vocabulary expansion for SeaLLM v1 and v2 with Llama-2 and Mistral-7B as backbones due to their limit 32K-token vocabulary. However, we did not extend the tokenizer for SeaLLM-7B-v2.5, which inherits a sufficiently large 250K-token vocabulary from Gemma-7B.

\subsection{Pre-training Process}\label{sec:pretrain:process}

We organize our pre-training dataset based on the language of the content and the quality of the data, as mentioned in \Cref{sec:pretrain:data}. We setup a separate stream of data for each language, and dynamically control and balance the sampling ratio of each language. We pack multilingual documents into a single sequence up to the maximum context length. During the last steps of pre-training, we re-feed the model with more high quality data, which it has previously seen, to readjust the model's learning focus back towards the high-quality data, improving the model's performance.

\section{Supervised Fine-tuning (SFT)}\label{sec:sft}

\subsection{Supervised Fine-tuning Data}\label{sec:sft:data}



Our supervised finetuning (SFT) data consists of many categories, including text understanding and processing, math and logical reasoning, user-centric instruction-following, and natural dialog data. 
As most public and open-source SFT data are English-only~\citep{flan_collection_longpre2023,openorca,orca_mukherjee2023orca,platypus2023}, 
various techniques were implemented to enhance the multilingual aspect of the model. These include sourcing natural data from local websites in natural settings, selectively translating from English data, employing self-instruction, and using advanced prompting techniques~\citep{self_instruct_wang2022self,self_refine_madaan2023self,democratize_llm_ldp_nguyen2023}. As those synthetically generated data may remain incorrect or low-quality, native speakers\footnote{Hired by our organization, they are not co-authors.} were then engaged to further verify, filter, and edit such synthetic responses to finalize the SFT dataset. We find that engaging the annotators to verify and modify model-generated responses is more efficient than having them write responses from scratch.
Safety-related data also played a crucial role in fine-tuning SeaLLMs.
We manually collected and prepared country-relevant safety data, which covered a broad range of culturally and legally sensitive topics in each of these countries. This was necessary as such topics are often overlooked or may even conflict with open-source English-centric safety data \cite{multilingual_jailbreak_deng2023}. 

For SeaLLM-7B-v2 and SeaLLM-7B-v2.5, we incorporate significantly more SFT data relating to math and commonsense reasoning. Such data is synthetically generated with SeaLLM-13B-v1, as well as strong English models \citep{jiang2024mixtral,qwen_bai2023} using a combination of few-shot paraphrasing and translation techniques \citep{yu2023metamath}.


\subsection{Supervised Fine-tuning}\label{sec:sft:pretrain_sft}


\paragraph{Pre-train and SFT Hybrid.} As our SFT data is still significantly English due to contributions of open-source data, directly conducting SFT on it may overshadow the smaller SEA language datasets. Therefore, we propose incorporating an additional step prior to complete fine-tuning, namely Pre-train \& SFT Hybrid. In this step, the model is further trained on a combination of the pre-training corpus and a large portion of English SFT data, leaving the remaining and more balanced amount of English SFT data to the next stage.
During this hybrid stage, the model processes both general pre-training content and instruction-following examples. We mask the source side of the instruction or supervised data to prevent the model from overfitting to the training examples and to reduce the risk of it simply memorizing the input data instead of learning the more generalized ability to follow instructions.




\paragraph{Supervised Fine-tuning.} We conduct supervised fine-tuning by compiling instructions from a variety of sources explained in \Cref{sec:sft:data}, combining them at random into a single, consolidated sequence to maximize efficiency. 
To enhance the multi-turn conversation capability, in the later stage of fine-tuning, we further artificially create multi-turn conversations by randomly joining several single-turn instructions together. 







\subsection{Self-Preferencing Optimization}\label{sec:dpo}



Alignment from human feedback preference has been key to the success of many AI-assistant language models \citep{rlhf_stiennon2020learning,llama2touvron2023llama,dpo_rafailov2023direct,instructgpt_ouyang2022training}. To save the cost of human preference annotation work, some have sought to use powerful LLMs like GPT-4 \citep{gpt4} to play the part of a preference data generator \citep{zephyr7b_tunstall2023}. However, that may not even be feasible for low-resource non-Latin languages because of the unfavorable tokenization of ChatGPT as explained in \Cref{sec:pretrain:vocab_extend}. In other words, even short prompts would exceed their context-length and the API-call costs would explode by up to 17 times.


Therefore, we use our own SeaLLM SFT models to generate preference data by asking it to indicate its preference between two of its own responses, given a question based on certain human-written criteria. To eliminate position bias, we swap the order of the responses and remove samples with inconsistent preference. The data is later used to employ direct preference optimization \citep{dpo_rafailov2023direct} to significantly improve the model abilities as an assistant. As such, unlike other works \citep{orca_mukherjee2023orca,zephyr7b_tunstall2023}, our models are free from relying on powerful close-sourced models like GPT-4 to improve the performance in low-resource languages. Our self-preferencing method also shares certain flavors with another self-rewarding mechanism \citep{yuan2024self_reward_lm}.\footnote{Our work was publicly available before \citet{yuan2024self_reward_lm}.}

\section{Evaluation}\label{sec:eval}

\subsection{Model Variants}\label{sec:eval:variants}

We trained multiple variants of SeaLLMs, as specified in the following.

\begin{itemize}
    \item \textbf{SeaLLM-7B-v1}: Trained from Llama-2-7B, it supports the 10 official languages used in Southeast Asia.
    \item \textbf{SeaLLM-13B-v1}: Trained from Llama-2-13B, it outperforms ChatGPT-3.5 in most non-Latin SEA languages (Khm, Lao, Mya and Tha) by large margins.
    \item \textbf{SeaLLM-7B-v2}: Trained from Mistral-7B, it outperforms SeaLLM-13B-v1 by far in higher-resource SEA languages (Vie, Ind, Tha), and surpasses ChatGPT-3.5 in math reasoning in SEA languages.
    \item \textbf{SeaLLM-7B-v2.5}: Trained from Gemma-7B, it outperforms SeaLLM-7B-v2 and SeaLLM-13B-v1 remarkably and surpasses ChatGPT-3.5 in various aspects in SEA languages, especially non-Latin languages.
\end{itemize}

\subsection{Sea-bench Peer Comparison}\label{sec:eval:peer_com}


\begin{table*}[h]
  \centering
  \resizebox{0.6\textwidth}{!}{%
  \begin{tabular}{ccccccc}
    \toprule
    \multirow{2}{*}{\bf Model} & \multicolumn{5}{c}{\bf M3Exam} & {\bf MMLU} \\
    \cmidrule{2-7}
     & Eng & Zho & Vie & Ind & Tha & Eng  \\
    \midrule
    ChatGPT-3.5         & 75.46 & 60.20 & 58.64 & \textbf{49.27} & 37.41 & \textbf{70.00} \\
    \midrule
    SeaLion-7b          & 23.80 & 25.87 & 27.11 & 24.28 & 20.29 & 26.87 \\
    Llama-2-13b         & 61.17 & 43.29 & 39.97 & 35.50 & 23.74 & 53.50 \\
    Polylm-13b          & 32.23 & 29.26 & 29.01 & 25.36 & 18.08 & 22.94 \\
    \midrule
    SeaLLM-7B-v1        & 54.89 & 39.30 & 38.74 & 32.95 & 25.09 & 47.16 \\
    SeaLLM-13B-v1        & 62.69 & 44.50 & 46.45 & 39.28 & 36.39 & 52.68 \\
    SeaLLM-7B-v2         & 70.91 & 55.43 & 51.15 & 42.25 & 35.52 & 61.89 \\
    SeaLLM-7B-v2.5       & \textbf{76.87} & \textbf{62.54} & \textbf{63.11} & 48.64 & \textbf{46.86} & 64.05 \\
    \bottomrule
  \end{tabular}
  }
  \caption{Multilingual world knowledge accuracy evaluation across multiple languages and various models of different sizes.}
  \label{table:eval:world_knowledge}  
\end{table*}

While there are popular benchmarks to evaluate LLMs as a helpful assistant, such as MT-bench \citep{mt_bench_zheng2023judging}, they are only English-based and not suitable for evaluating performances in low-resource languages. Due to such a lack of multilingual benchmarks for assistant-style models, we engaged native linguists to build a multilingual test set with instructions that cover SEA languages, called \textbf{Sea-bench}. The linguists sourced such data by translating open-source English test sets, collecting real user questions from local forums and websites, collecting real math and reasoning questions from reputable sources, as well as writing test instructions themselves.
Our Sea-Bench consists of diverse categories of instructions to evaluate the models, as follows:
\begin{itemize}
    \item Task-solving: This type of data comprises various text understanding and processing tasks, such as summarization, translation, etc.
    \item Math-reasoning: This includes math problems and logical reasoning tasks.
    \item General-instruction data: This consists of general user-centric instructions, which evaluate the model's ability in general knowledge and writing.
    \item NaturalQA: This consists of queries posted by real users, often in popular local forums, with a variety of subjects and topics of local interest. The aim is to test the model’s capacity to understand and respond coherently to colloquial language, natural expressions and idiomatic language, and locally contextualized references.
    \item Safety: This includes both general safety and local context-related safety instructions. While most general safety questions are translated from open sources, other local country-specific safety instructions are written by linguists of each language.
\end{itemize}

As inspired by MT-bench \citep{mt_bench_zheng2023judging}, we evaluate and compare SeaLLMs with well-known and state-of-the-art models using GPT-4 as a judge in a score-based grading metrics and a peer comparison (or pairwise comparison) manner. 

\Cref{fig:sea_bench:seallm_vs_chatgpt} compares our SeaLLM (v2, v2.5) chat models with Qwen1.5-7B-chat \citep{qwen_bai2023} and the widely reputed ChatGPT-3.5\footnote{gpt-3.5-turbo June 2023 version.} \citep{chatgpt}. In the ``By Category'' chart, SeaLLM-7B-v2.5 performs on par with or surpasses ChatGPT-3.5 across various linguistic and writing tasks. This is largely thanks to the large gap in low-resource non-Latin languages, such as Burmese (Mya), Lao, Khmer and Thai, as seen in the ``By language'' chart on the right in \Cref{fig:sea_bench:seallm_vs_chatgpt}. 

\begin{table}[h]
  \centering
  \resizebox{\columnwidth}{!}{%
  \begin{tabular}{cccc}
    \toprule
    {\bf Model} & {\bf Languages} & {\bf MT-bench} \\
    \midrule
    GPT-4-turbo     & Multi & 9.32 \\
    \midrule
    Mixtral-8x7B (46B)      & Multi & 8.3 \\
    Starling-LM-7B-alpha    & Mono (Eng) & 8.0 \\
    OpenChat-3.5-7B         & Mono (Eng) & 7.81 \\
    \textbf{SeaLLM-7B-v2}   & \textbf{Multi}      & \textbf{7.54} \\
    \textbf{SeaLLM-7B-v2.5}   & \textbf{Multi}      & \textbf{7.43} \\
    Llama-2-70B-chat        & Mono       & 6.86 \\
    Mistral-7B-instruct     & Mono       & 6.84 \\
    \textbf{SeaLLM-13B-v1}  & \textbf{Multi}      & \textbf{6.32} \\
    \bottomrule
  \end{tabular}
  }
  \caption{MT-Bench scores \citep{mt_bench_zheng2023judging} for closed, open, multilingual and monolingual (as indicated by their authors on Huggingface.) models.}
  \label{table:eval:mt_bench}
\end{table}


\begin{table*}[h]
  \centering
  \resizebox{\textwidth}{!}{%
  \begin{tabular}{ccccccccccc}
    \toprule
    \multirow{2}{*}{\bf Model}	& \multicolumn{2}{c}{\bf Eng}	& \multicolumn{2}{c}{\bf Zho}	& \multicolumn{2}{c}{\bf Vie}	& \multicolumn{2}{c}{\bf Ind}	& \multicolumn{2}{c}{\bf Tha} \\
        & GSM8K & MATH & GSM8K & MATH & GSM8K & MATH & GSM8K & MATH & GSM8K & MATH \\
    \midrule
    ChatGPT-3.5     & \textbf{80.8} & 34.1 & 48.2 & 21.5 & 55.0 & 26.5 & 64.3 & 26.4 & 35.8 & 18.1 \\
    Qwen1.5-7B-chat & 56.8 & 15.3 & 40.0 & 2.7 & 37.7 & 9.0 & 36.9 & 7.7 & 21.9 & 4.7 \\
    SeaLLM-7B-v2    & 78.2 & 27.5 & \textbf{53.7} & 17.6 & 69.9 & 23.8 & \textbf{71.5} & 24.4 & 59.6 & 22.4 \\
    SeaLLM-7B-v2.5  & 78.5 & \textbf{34.9} & 51.3 & \textbf{22.1} & \textbf{72.3} & \textbf{30.2} & \textbf{71.5} & \textbf{30.1} & \textbf{62.0} & \textbf{28.4} \\
    \bottomrule
  \end{tabular}
  }
  \caption{GSM8K and MATH scores \citep{gsm8k_cobbe2021gsm8k,MATH_hendrycksmath2021} and their translated-versions in Chinese, Vietnamese, Indonesian and Thai, under zero-shot chain-of-thought prompting for different models.}
  \label{table:eval:sea_math}
\end{table*}

\subsection{MT-bench}\label{sec:eval:mt_bench}

We also compare our models with certain baselines on the English MT-Bench \citep{mt_bench_zheng2023judging} in \Cref{table:eval:mt_bench}. As shown, SeaLLM-7B-v2 model demonstrates outstanding ability in English, given its size. It is also a rare multilingual model in the 7B realm, especially since it focuses on non-mainstream languages.

\begin{figure}[t] 
    \centering
    \includegraphics[width=\columnwidth]{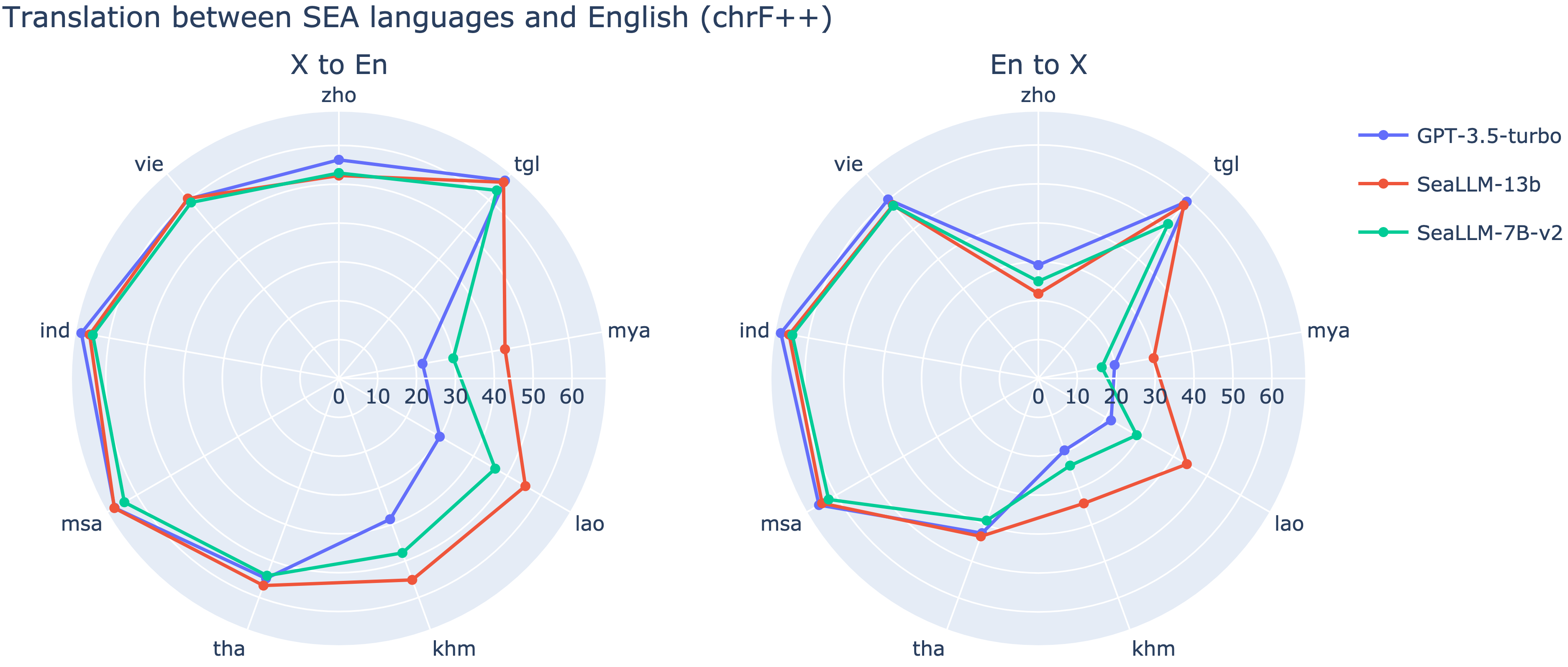} 
    \caption{Translation chrF++ scores of various models for both SEA languages to English and English to SEA languages directions.}
    \label{fig:translation}
\end{figure}

\begin{figure}[t] 
    \centering
    \includegraphics[width=\columnwidth]{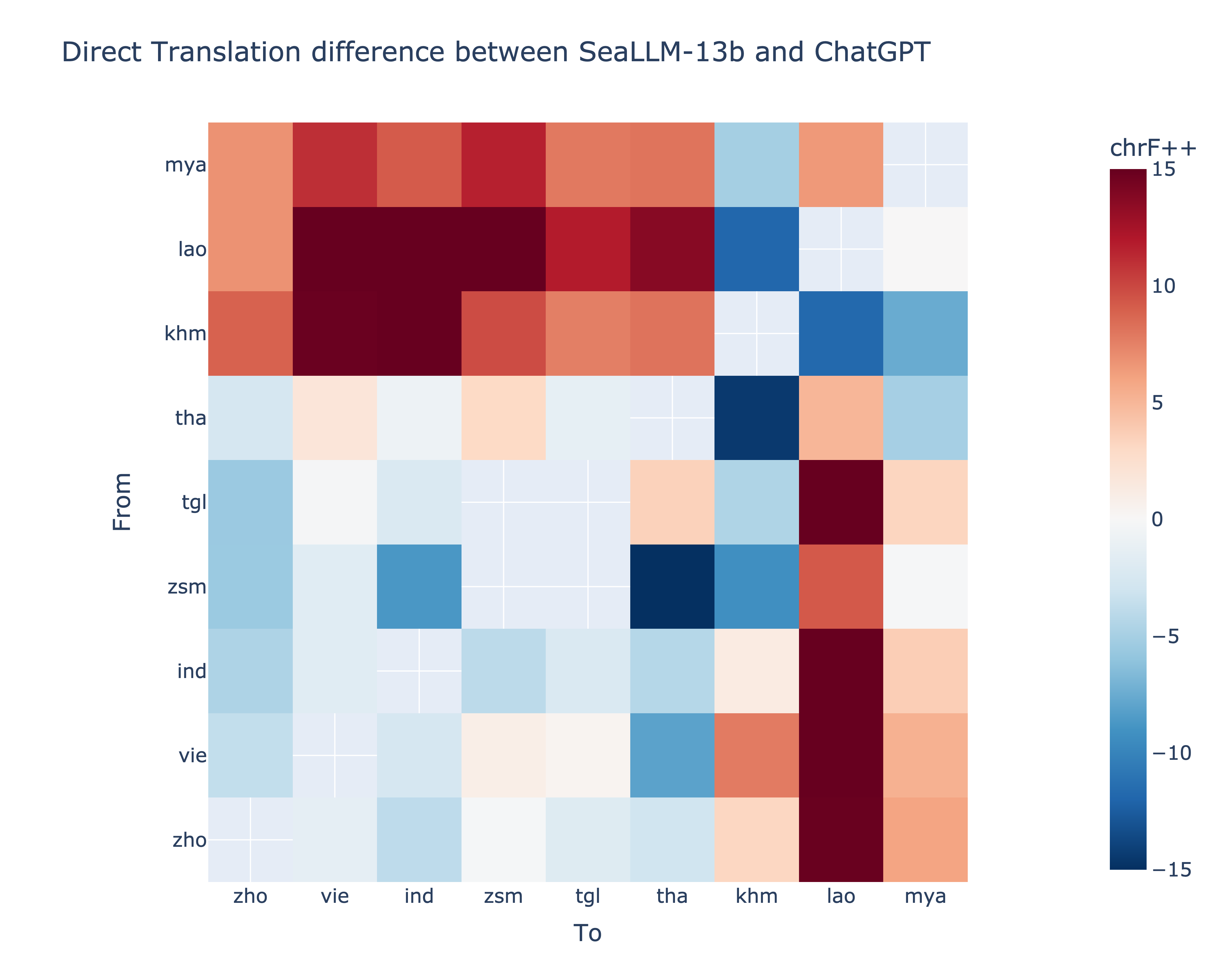} 
    \caption{Direct translation between SEA languages. Scores are indicated as the different between the respective chrF++ score of SeaLLM-13B-v1 minus that of ChatGPT-3.5. Red colors suggests SeaLLM-13B-v1 is better, while blue colors indicates ChatGPT is better.}
    \label{fig:direct_translation}
\end{figure}

\subsection{World Knowledge}\label{sec:eval:knowledge}


In this section, we evaluate our models and reputable chat baselines \citep{llama2touvron2023llama,polylm_wei2023polylm,chatgpt} in terms of world knowledge. For knowledge across languages, we use the M3Exam benchmark \citep{m3exam}, which consists of real questions from human exam papers with various degrees of difficulty, ranging from primary school to high school examinations. We evaluate M3Exam with 3-shot native-instruction prompts across English, Chinese, Vietnamese, Indonesian and Thai. We also evaluate our models with the well-known English-centric MMLU benchmark \citep{mmlu_hendryckstest2021}.

\Cref{table:eval:world_knowledge} details the evaluations of world knowledge across multiple languages and models of different sizes. SeaLLM-7B-v2.5 exhibits the best performance given its size and is competitive to GPT-3.5.

\subsection{Math Reasoning}\label{sec:eval:math_reasoning}

\Cref{table:eval:sea_math} shows the GSM8K and MATH scores \citep{gsm8k_cobbe2021gsm8k,MATH_hendrycksmath2021} for zero-shot chain-of-thought prompting for English and their translated version in Chinese, Vietnamese, Indonesian and Thai. As shown, SeaLLM-7B-v2.5 shows competitive English performance in math reasoning compared to open-source models, with 78.5 in GSM8K and 34.9 in MATH. It also exceeds GPT-3.5 in SEA languages. This is achieved by scaling supervised and preference data in math reasoning in multilingual settings.

\subsection{Machine Translation}\label{sec:eval:translation}
To benchmark the machine translation performance of our SeaLLMs, we evaluate 4-shot chrF++ scores on the test sets from Flores-200~\cite{nllb_costa2022no_flores200}. As can be seen from \Cref{fig:translation}, SeaLLM-13B exhibits clear superiority over ChatGPT-3.5 in low-resource languages, such as Lao and Khmer, while maintaining comparable performance with ChatGPT-3.5 in most higher resource languages (e.g., Vietnamese and Indonesian). 

For direct translation between SEA languages, as shown in \Cref{fig:direct_translation}, our SeaLLM-13B-v1 model still achieves higher chrF++ scores than ChatGPT-3.5 in most cases, especially when the translation pairs involve low-resource languages. Overall, we believe our SeaLLMs will play a key role in facilitating communication and cultural exchange across communities in Southeast Asia.



\section{Conclusion}
In conclusion, our research presents a substantial advance in the development of equitable and culturally aware AI with the creation of SeaLLMs, a specialized suite of language models attuned to the linguistic and cultural landscapes of Southeast Asia. Through rigorous pre-training enhancements and culturally tailored fine-tuning processes, SeaLLMs have demonstrated exceptional proficiency in language understanding and generation tasks, challenging the performance of dominant players such as ChatGPT-3.5, particularly in SEA languages. The models' attunement to local norms and legal stipulations—validated by human evaluations—establishes SeaLLMs as not only a technical breakthrough but a socially responsive innovation, poised to democratize access to high-quality AI language tools across linguistically diverse regions. This work lays a foundation for further research into language models that respect and uphold the rich tapestry of human languages and cultures, ultimately driving the AI community towards a more inclusive future.

\section{Limitations}

SeaLLMs are among the most linguistically diverse multilingual large language models with remarkable abilities in languages beyond mainstream. However, they do not come without limitations. First, they only scratch the surface of the regionally linguistic diversity with 9 most common and representative languages, while there are hundreds other languages spoken in the Southeast Asia, such as Javanese and Tamil. Second, despite outperforming other popular models in non-Latin low-resource languages, SeaLLM models still suffer from considerable hallucination and degeneration under certain circumstances for languages such as Burmese and Lao. Moderate hallucination is still inevitable for other common languages.

\section{Acknowledgement}

We would like to express our special thanks to our professional and native linguists, Tantong Champaiboon, Nguyen Ngoc Yen Nhi and Tara Devina Putri, who helped build, evaluate, and fact-check our sampled pretraining and SFT dataset as well as evaluating our models across different aspects, especially safety.

\bibliography{anthology,acl_latex}

\begin{thebibliography}{42}
\expandafter\ifx\csname natexlab\endcsname\relax\def\natexlab#1{#1}\fi

\bibitem[{Ahuja et~al.(2023)Ahuja, Hada, Ochieng, Jain, Diddee, Maina, Ganu, Segal, Axmed, Bali, and Sitaram}]{mega}
Kabir Ahuja, Rishav Hada, Millicent Ochieng, Prachi Jain, Harshita Diddee, Samuel Maina, Tanuja Ganu, Sameer Segal, Maxamed Axmed, Kalika Bali, and Sunayana Sitaram. 2023.
\newblock {MEGA:} multilingual evaluation of generative {AI}.
\newblock \emph{CoRR}, abs/2303.12528.

\bibitem[{Bai et~al.(2023)Bai, Bai, Chu, Cui, Dang, Deng, Fan, Ge, Han, Huang et~al.}]{qwen_bai2023}
Jinze Bai, Shuai Bai, Yunfei Chu, Zeyu Cui, Kai Dang, Xiaodong Deng, Yang Fan, Wenbin Ge, Yu~Han, Fei Huang, et~al. 2023.
\newblock Qwen technical report.
\newblock \emph{arXiv preprint arXiv:2309.16609}.

\bibitem[{Bojanowski et~al.(2017)Bojanowski, Grave, Joulin, and Mikolov}]{fasttext}
Piotr Bojanowski, Edouard Grave, Armand Joulin, and Tomas Mikolov. 2017.
\newblock Enriching word vectors with subword information.
\newblock \emph{Transactions of the Association for Computational Linguistics}, 5:135--146.

\bibitem[{Brown et~al.(2020)Brown, Mann, Ryder, Subbiah, Kaplan, Dhariwal, Neelakantan, Shyam, Sastry, Askell et~al.}]{gpt3_brown2020language}
Tom Brown, Benjamin Mann, Nick Ryder, Melanie Subbiah, Jared~D Kaplan, Prafulla Dhariwal, Arvind Neelakantan, Pranav Shyam, Girish Sastry, Amanda Askell, et~al. 2020.
\newblock Language models are few-shot learners.
\newblock \emph{Advances in neural information processing systems}, 33:1877--1901.

\bibitem[{Cobbe et~al.(2021)Cobbe, Kosaraju, Bavarian, Chen, Jun, Kaiser, Plappert, Tworek, Hilton, Nakano, Hesse, and Schulman}]{gsm8k_cobbe2021gsm8k}
Karl Cobbe, Vineet Kosaraju, Mohammad Bavarian, Mark Chen, Heewoo Jun, Lukasz Kaiser, Matthias Plappert, Jerry Tworek, Jacob Hilton, Reiichiro Nakano, Christopher Hesse, and John Schulman. 2021.
\newblock Training verifiers to solve math word problems.
\newblock \emph{arXiv preprint arXiv:2110.14168}.

\bibitem[{Computer(2023)}]{together2023redpajama}
Together Computer. 2023.
\newblock \href {https://github.com/togethercomputer/RedPajama-Data} {Redpajama: An open source recipe to reproduce llama training dataset}.

\bibitem[{Costa-juss{\`a} et~al.(2022)Costa-juss{\`a}, Cross, {\c{C}}elebi, Elbayad, Heafield, Heffernan, Kalbassi, Lam, Licht, Maillard et~al.}]{nllb_costa2022no_flores200}
Marta~R Costa-juss{\`a}, James Cross, Onur {\c{C}}elebi, Maha Elbayad, Kenneth Heafield, Kevin Heffernan, Elahe Kalbassi, Janice Lam, Daniel Licht, Jean Maillard, et~al. 2022.
\newblock No language left behind: Scaling human-centered machine translation.
\newblock \emph{arXiv preprint arXiv:2207.04672}.

\bibitem[{Deng et~al.(2023)Deng, Zhang, Pan, and Bing}]{multilingual_jailbreak_deng2023}
Yue Deng, Wenxuan Zhang, Sinno~Jialin Pan, and Lidong Bing. 2023.
\newblock Multilingual jailbreak challenges in large language models.
\newblock \emph{arXiv preprint arXiv:2310.06474}.

\bibitem[{Foundation()}]{wikidump}
Wikimedia Foundation.
\newblock \href {https://dumps.wikimedia.org} {Wikimedia downloads}.

\bibitem[{Hendrycks et~al.(2021{\natexlab{a}})Hendrycks, Burns, Basart, Zou, Mazeika, Song, and Steinhardt}]{mmlu_hendryckstest2021}
Dan Hendrycks, Collin Burns, Steven Basart, Andy Zou, Mantas Mazeika, Dawn Song, and Jacob Steinhardt. 2021{\natexlab{a}}.
\newblock Measuring massive multitask language understanding.
\newblock \emph{Proceedings of the International Conference on Learning Representations (ICLR)}.

\bibitem[{Hendrycks et~al.(2021{\natexlab{b}})Hendrycks, Burns, Kadavath, Arora, Basart, Tang, Song, and Steinhardt}]{MATH_hendrycksmath2021}
Dan Hendrycks, Collin Burns, Saurav Kadavath, Akul Arora, Steven Basart, Eric Tang, Dawn Song, and Jacob Steinhardt. 2021{\natexlab{b}}.
\newblock Measuring mathematical problem solving with the math dataset.
\newblock \emph{NeurIPS}.

\bibitem[{Jiang et~al.(2023)Jiang, Sablayrolles, Mensch, Bamford, Chaplot, Casas, Bressand, Lengyel, Lample, Saulnier et~al.}]{mistral7b_jiang2023mistral}
Albert~Q Jiang, Alexandre Sablayrolles, Arthur Mensch, Chris Bamford, Devendra~Singh Chaplot, Diego de~las Casas, Florian Bressand, Gianna Lengyel, Guillaume Lample, Lucile Saulnier, et~al. 2023.
\newblock Mistral 7b.
\newblock \emph{arXiv preprint arXiv:2310.06825}.

\bibitem[{Jiang et~al.(2024)Jiang, Sablayrolles, Roux, Mensch, Savary, Bamford, Chaplot, Casas, Hanna, Bressand et~al.}]{jiang2024mixtral}
Albert~Q Jiang, Alexandre Sablayrolles, Antoine Roux, Arthur Mensch, Blanche Savary, Chris Bamford, Devendra~Singh Chaplot, Diego de~las Casas, Emma~Bou Hanna, Florian Bressand, et~al. 2024.
\newblock Mixtral of experts.
\newblock \emph{arXiv preprint arXiv:2401.04088}.

\bibitem[{Lai et~al.(2023)Lai, Ngo, Veyseh, Man, Dernoncourt, Bui, and Nguyen}]{multilingual-eval}
Viet~Dac Lai, Nghia~Trung Ngo, Amir Pouran~Ben Veyseh, Hieu Man, Franck Dernoncourt, Trung Bui, and Thien~Huu Nguyen. 2023.
\newblock Chatgpt beyond english: Towards a comprehensive evaluation of large language models in multilingual learning.
\newblock \emph{CoRR}, abs/2304.05613.

\bibitem[{Lee et~al.(2023)Lee, Hunter, and Ruiz}]{platypus2023}
Ariel~N. Lee, Cole~J. Hunter, and Nataniel Ruiz. 2023.
\newblock Platypus: Quick, cheap, and powerful refinement of llms.

\bibitem[{Lian et~al.(2023)Lian, Goodson, Pentland, Cook, Vong, and "Teknium"}]{openorca}
Wing Lian, Bleys Goodson, Eugene Pentland, Austin Cook, Chanvichet Vong, and "Teknium". 2023.
\newblock Openorca: An open dataset of gpt augmented flan reasoning traces.
\newblock \url{https://https://huggingface.co/Open-Orca/OpenOrca}.

\bibitem[{Longpre et~al.(2023)Longpre, Hou, Vu, Webson, Chung, Tay, Zhou, Le, Zoph, Wei, and Roberts}]{flan_collection_longpre2023}
Shayne Longpre, Le~Hou, Tu~Vu, Albert Webson, Hyung~Won Chung, Yi~Tay, Denny Zhou, Quoc~V. Le, Barret Zoph, Jason Wei, and Adam Roberts. 2023.
\newblock \href {http://arxiv.org/abs/2301.13688} {The flan collection: Designing data and methods for effective instruction tuning}.

\bibitem[{Madaan et~al.(2023)Madaan, Tandon, Gupta, Hallinan, Gao, Wiegreffe, Alon, Dziri, Prabhumoye, Yang et~al.}]{self_refine_madaan2023self}
Aman Madaan, Niket Tandon, Prakhar Gupta, Skyler Hallinan, Luyu Gao, Sarah Wiegreffe, Uri Alon, Nouha Dziri, Shrimai Prabhumoye, Yiming Yang, et~al. 2023.
\newblock Self-refine: Iterative refinement with self-feedback.
\newblock \emph{arXiv preprint arXiv:2303.17651}.

\bibitem[{Muennighoff et~al.(2022)Muennighoff, Wang, Sutawika, Roberts, Biderman, Scao, Bari, Shen, Yong, Schoelkopf et~al.}]{bloomz_muennighoff2022crosslingual}
Niklas Muennighoff, Thomas Wang, Lintang Sutawika, Adam Roberts, Stella Biderman, Teven~Le Scao, M~Saiful Bari, Sheng Shen, Zheng-Xin Yong, Hailey Schoelkopf, et~al. 2022.
\newblock Crosslingual generalization through multitask finetuning.
\newblock \emph{arXiv preprint arXiv:2211.01786}.

\bibitem[{Mukherjee et~al.(2023)Mukherjee, Mitra, Jawahar, Agarwal, Palangi, and Awadallah}]{orca_mukherjee2023orca}
Subhabrata Mukherjee, Arindam Mitra, Ganesh Jawahar, Sahaj Agarwal, Hamid Palangi, and Ahmed Awadallah. 2023.
\newblock \href {http://arxiv.org/abs/2306.02707} {Orca: Progressive learning from complex explanation traces of gpt-4}.

\bibitem[{Nguyen et~al.(2023)Nguyen, Aljunied, Joty, and Bing}]{democratize_llm_ldp_nguyen2023}
Xuan-Phi Nguyen, Sharifah~Mahani Aljunied, Shafiq Joty, and Lidong Bing. 2023.
\newblock Democratizing llms for low-resource languages by leveraging their english dominant abilities with linguistically-diverse prompts.
\newblock \emph{arXiv preprint arXiv:2306.11372}.

\bibitem[{OpenAI(2023{\natexlab{a}})}]{chatgpt}
OpenAI. 2023{\natexlab{a}}.
\newblock Chatgpt (june 2023 version.

\bibitem[{OpenAI(2023{\natexlab{b}})}]{gpt4}
OpenAI. 2023{\natexlab{b}}.
\newblock Gpt-4 technical report.
\newblock \emph{arXiv preprint}.

\bibitem[{Ouyang et~al.(2022)Ouyang, Wu, Jiang, Almeida, Wainwright, Mishkin, Zhang, Agarwal, Slama, Ray et~al.}]{instructgpt_ouyang2022training}
Long Ouyang, Jeffrey Wu, Xu~Jiang, Diogo Almeida, Carroll Wainwright, Pamela Mishkin, Chong Zhang, Sandhini Agarwal, Katarina Slama, Alex Ray, et~al. 2022.
\newblock Training language models to follow instructions with human feedback.
\newblock \emph{Advances in Neural Information Processing Systems}, 35:27730--27744.

\bibitem[{Rafailov et~al.(2023)Rafailov, Sharma, Mitchell, Ermon, Manning, and Finn}]{dpo_rafailov2023direct}
Rafael Rafailov, Archit Sharma, Eric Mitchell, Stefano Ermon, Christopher~D Manning, and Chelsea Finn. 2023.
\newblock Direct preference optimization: Your language model is secretly a reward model.
\newblock \emph{arXiv preprint arXiv:2305.18290}.

\bibitem[{Raffel et~al.(2019)Raffel, Shazeer, Roberts, Lee, Narang, Matena, Zhou, Li, and Liu}]{mc4_2019t5}
Colin Raffel, Noam Shazeer, Adam Roberts, Katherine Lee, Sharan Narang, Michael Matena, Yanqi Zhou, Wei Li, and Peter~J. Liu. 2019.
\newblock \href {http://arxiv.org/abs/1910.10683} {Exploring the limits of transfer learning with a unified text-to-text transformer}.
\newblock \emph{arXiv e-prints}.

\bibitem[{Scao et~al.(2022)Scao, Fan, Akiki, Pavlick, Ili{\'c}, Hesslow, Castagn{\'e}, Luccioni, Yvon, Gall{\'e} et~al.}]{bloom_scao2022bloom}
Teven~Le Scao, Angela Fan, Christopher Akiki, Ellie Pavlick, Suzana Ili{\'c}, Daniel Hesslow, Roman Castagn{\'e}, Alexandra~Sasha Luccioni, Fran{\c{c}}ois Yvon, Matthias Gall{\'e}, et~al. 2022.
\newblock Bloom: A 176b-parameter open-access multilingual language model.
\newblock \emph{arXiv preprint arXiv:2211.05100}.

\bibitem[{Sennrich et~al.(2016)Sennrich, Haddow, and Birch}]{bpe_sennrich2015neural}
Rico Sennrich, Barry Haddow, and Alexandra Birch. 2016.
\newblock \href {https://doi.org/10.18653/v1/P16-1162} {Neural machine translation of rare words with subword units}.
\newblock In \emph{Proceedings of the 54th Annual Meeting of the Association for Computational Linguistics (Volume 1: Long Papers)}, pages 1715--1725. Association for Computational Linguistics.

\bibitem[{Stiennon et~al.(2020)Stiennon, Ouyang, Wu, Ziegler, Lowe, Voss, Radford, Amodei, and Christiano}]{rlhf_stiennon2020learning}
Nisan Stiennon, Long Ouyang, Jeffrey Wu, Daniel Ziegler, Ryan Lowe, Chelsea Voss, Alec Radford, Dario Amodei, and Paul~F Christiano. 2020.
\newblock Learning to summarize with human feedback.
\newblock \emph{Advances in Neural Information Processing Systems}, 33:3008--3021.

\bibitem[{Su{\'a}rez et~al.(2019)Su{\'a}rez, Sagot, and Romary}]{suarez2019asynchronous}
Pedro Javier~Ortiz Su{\'a}rez, Beno{\^\i}t Sagot, and Laurent Romary. 2019.
\newblock Asynchronous pipeline for processing huge corpora on medium to low resource infrastructures.
\newblock In \emph{7th Workshop on the Challenges in the Management of Large Corpora (CMLC-7)}. Leibniz-Institut f{\"u}r Deutsche Sprache.

\bibitem[{Team et~al.(2024)Team, Mesnard, Hardin, Dadashi, Bhupatiraju, Pathak, Sifre, Rivi{\`e}re, Kale, Love et~al.}]{gemma_team2024}
Gemma Team, Thomas Mesnard, Cassidy Hardin, Robert Dadashi, Surya Bhupatiraju, Shreya Pathak, Laurent Sifre, Morgane Rivi{\`e}re, Mihir~Sanjay Kale, Juliette Love, et~al. 2024.
\newblock Gemma: Open models based on gemini research and technology.
\newblock \emph{arXiv preprint arXiv:2403.08295}.

\bibitem[{Thoppilan et~al.(2022)Thoppilan, De~Freitas, Hall, Shazeer, Kulshreshtha, Cheng, Jin, Bos, Baker, Du et~al.}]{lambda_google_thoppilan2022lamda}
Romal Thoppilan, Daniel De~Freitas, Jamie Hall, Noam Shazeer, Apoorv Kulshreshtha, Heng-Tze Cheng, Alicia Jin, Taylor Bos, Leslie Baker, Yu~Du, et~al. 2022.
\newblock Lamda: Language models for dialog applications.
\newblock \emph{arXiv preprint arXiv:2201.08239}.

\bibitem[{Touvron et~al.(2023{\natexlab{a}})Touvron, Lavril, Izacard, Martinet, Lachaux, Lacroix, Rozi{\`e}re, Goyal, Hambro, Azhar, Rodriguez, Joulin, Grave, and Lample}]{llama_touvron2023}
Hugo Touvron, Thibaut Lavril, Gautier Izacard, Xavier Martinet, Marie-Anne Lachaux, Timoth{\'e}e Lacroix, Baptiste Rozi{\`e}re, Naman Goyal, Eric Hambro, Faisal Azhar, Aurelien Rodriguez, Armand Joulin, Edouard Grave, and Guillaume Lample. 2023{\natexlab{a}}.
\newblock Llama: Open and efficient foundation language models.
\newblock \emph{arXiv preprint arXiv:2302.13971}.

\bibitem[{Touvron et~al.(2023{\natexlab{b}})Touvron, Martin, Stone, Albert, Almahairi, Babaei, Bashlykov, Batra, Bhargava, Bhosale et~al.}]{llama2touvron2023llama}
Hugo Touvron, Louis Martin, Kevin Stone, Peter Albert, Amjad Almahairi, Yasmine Babaei, Nikolay Bashlykov, Soumya Batra, Prajjwal Bhargava, Shruti Bhosale, et~al. 2023{\natexlab{b}}.
\newblock Llama 2: Open foundation and fine-tuned chat models.
\newblock \emph{arXiv preprint arXiv:2307.09288}.

\bibitem[{Tunstall et~al.(2023)Tunstall, Beeching, Lambert, Rajani, Rasul, Belkada, Huang, von Werra, Fourrier, Habib, Sarrazin, Sanseviero, Rush, and Wolf}]{zephyr7b_tunstall2023}
Lewis Tunstall, Edward Beeching, Nathan Lambert, Nazneen Rajani, Kashif Rasul, Younes Belkada, Shengyi Huang, Leandro von Werra, Clémentine Fourrier, Nathan Habib, Nathan Sarrazin, Omar Sanseviero, Alexander~M. Rush, and Thomas Wolf. 2023.
\newblock \href {http://arxiv.org/abs/2310.16944} {Zephyr: Direct distillation of lm alignment}.

\bibitem[{Wang et~al.(2022)Wang, Kordi, Mishra, Liu, Smith, Khashabi, and Hajishirzi}]{self_instruct_wang2022self}
Yizhong Wang, Yeganeh Kordi, Swaroop Mishra, Alisa Liu, Noah~A Smith, Daniel Khashabi, and Hannaneh Hajishirzi. 2022.
\newblock Self-instruct: Aligning language model with self generated instructions.
\newblock \emph{arXiv preprint arXiv:2212.10560}.

\bibitem[{Wei et~al.(2023)Wei, Wei, Lin, Li, Zhang, Ren, Li, Wan, Cao, Xie et~al.}]{polylm_wei2023polylm}
Xiangpeng Wei, Haoran Wei, Huan Lin, Tianhao Li, Pei Zhang, Xingzhang Ren, Mei Li, Yu~Wan, Zhiwei Cao, Binbin Xie, et~al. 2023.
\newblock Polylm: An open source polyglot large language model.
\newblock \emph{arXiv preprint arXiv:2307.06018}.

\bibitem[{Wenzek et~al.(2020)Wenzek, Lachaux, Conneau, Chaudhary, Guzm{\'a}n, Joulin, and Grave}]{ccnet_wenzek-etal-2020-ccnet}
Guillaume Wenzek, Marie-Anne Lachaux, Alexis Conneau, Vishrav Chaudhary, Francisco Guzm{\'a}n, Armand Joulin, and Edouard Grave. 2020.
\newblock \href {https://www.aclweb.org/anthology/2020.lrec-1.494} {{CCN}et: Extracting high quality monolingual datasets from web crawl data}.
\newblock In \emph{Proceedings of the 12th Language Resources and Evaluation Conference}, pages 4003--4012, Marseille, France. European Language Resources Association.

\bibitem[{Yu et~al.(2023)Yu, Jiang, Shi, Yu, Liu, Zhang, Kwok, Li, Weller, and Liu}]{yu2023metamath}
Longhui Yu, Weisen Jiang, Han Shi, Jincheng Yu, Zhengying Liu, Yu~Zhang, James~T Kwok, Zhenguo Li, Adrian Weller, and Weiyang Liu. 2023.
\newblock Metamath: Bootstrap your own mathematical questions for large language models.
\newblock \emph{arXiv preprint arXiv:2309.12284}.

\bibitem[{Yuan et~al.(2024)Yuan, Pang, Cho, Sukhbaatar, Xu, and Weston}]{yuan2024self_reward_lm}
Weizhe Yuan, Richard~Yuanzhe Pang, Kyunghyun Cho, Sainbayar Sukhbaatar, Jing Xu, and Jason Weston. 2024.
\newblock Self-rewarding language models.
\newblock \emph{arXiv preprint arXiv:2401.10020}.

\bibitem[{Zhang et~al.(2023)Zhang, Aljunied, Gao, Chia, and Bing}]{m3exam}
Wenxuan Zhang, Sharifah~Mahani Aljunied, Chang Gao, Yew~Ken Chia, and Lidong Bing. 2023.
\newblock M3exam: {A} multilingual, multimodal, multilevel benchmark for examining large language models.
\newblock \emph{CoRR}, abs/2306.05179.

\bibitem[{Zheng et~al.(2023)Zheng, Chiang, Sheng, Zhuang, Wu, Zhuang, Lin, Li, Li, Xing et~al.}]{mt_bench_zheng2023judging}
Lianmin Zheng, Wei-Lin Chiang, Ying Sheng, Siyuan Zhuang, Zhanghao Wu, Yonghao Zhuang, Zi~Lin, Zhuohan Li, Dacheng Li, Eric Xing, et~al. 2023.
\newblock Judging llm-as-a-judge with mt-bench and chatbot arena.
\newblock \emph{arXiv preprint arXiv:2306.05685}.

\end{thebibliography}
\bibliographystyle{acl_natbib}

\appendix

\label{sec:appendix}


\section{Vocabulary Expansion}\label{app:vocab_extend}

\Cref{algo:vocab_extend} explains in details how we perform selective and recursive merger of tokens from target NLLB vocabulary into the original Llama vocabulary to enrich the linguistic coverage for new and low-resource languages. Specifically, given a small seed unlabeled dataset of a given new language, the algorithm first tokenizes a document with the current Llama tokenizer. The resulting tokens are then exhaustively merged into longer tokens that are supported by the target NLLB vocabulary. During this merger process, any intermediate sub-word is also added to the Llama tokenizer as long as they exist in the rich NLLB vocabulary.

The new set of collected tokens are then pruned to remove rarely appearing and low-quality tokens before being added to the final SeaLLM tokenizer. This frequency-based pruning process ensures the new language is sufficiently and efficiently encoded without introducing tokens from other existing languages (\eg\ English), which may disrupt the learned knowledge during the Llama-2 pre-training stage.



\begin{algorithm*}
\caption{Vocabulary Extension algorithm: $V_i$ is Llama vocabulary, $V_t$ is target NLLB vocabulary, $D$ is unlabeled data and $m$ is minimum frequency.}
\label{algo:vocab_extend}
\begin{algorithmic}[1]
\Function{ExhaustiveMerge}{$V_i, V_t, t_V$}
    \State $T_{new} \gets \text{empty set } \emptyset$
    \Repeat
        \For{each consecutive token pair $(\text{prev}, \text{next})$ in $t_V$}
            \State $t_{merged} \gets \langle \text{prev} \rangle \langle \text{next} \rangle$  \Comment{Form a new token}
            \If{$t_{merged}$ exists in $V_t$}
                \State Replace $(\text{prev}, \text{next})$ with $t_{merged}$ in  $t_V$ \Comment{Update $t_V$ with new token}
                \State $T_{new} \gets T_{new} \cup t_{merged}$
                \State \textbf{break}
            \EndIf
        \EndFor
    \Until{no new token added to $T_{new}$}
    \State \Return $T_{new}$
\EndFunction

\Function{VocabExtend}{$V_i, V_t, D, m$}
    \State $V \gets V_i$
    \State $F \gets \text{empty set } \emptyset$
    \State $T \gets \text{empty set } \emptyset$
    \For{document $d$ in $D$}
        \State $t_{V} \gets \text{tokenize}(V, d)$  \Comment{tokenize the document}
        \State $T_{new} \gets \Call{ExhaustiveMerge}{V_i, V_t, t_{V}}$ \Comment{obtain new words from $V_t$ based on $d$}
        \State $V \gets V \cup T_{new}$       \Comment{update $V$ with new words $T_{new}$}
        \State $T \gets T \cup T_{new}$
        \State $F \gets $ Update frequencies of $T_{new}$ to $F$ \Comment{update appearance frequencies of $T_{new}$}
    \EndFor
    \State $T \gets $ Prune $t_i \in T$ with corresponding $f_t \in F$ where $f_t < m$  \Comment{Remove rare words}
    \State $V_{final} \gets V_i \cup T$
    \State \Return $V_{final}$
\EndFunction
\end{algorithmic}
\end{algorithm*}

\section{Sea-bench Evaluation Details}\label{app:sea_bench}

\begin{figure*}[t] 
    \centering
    \includegraphics[width=\textwidth]{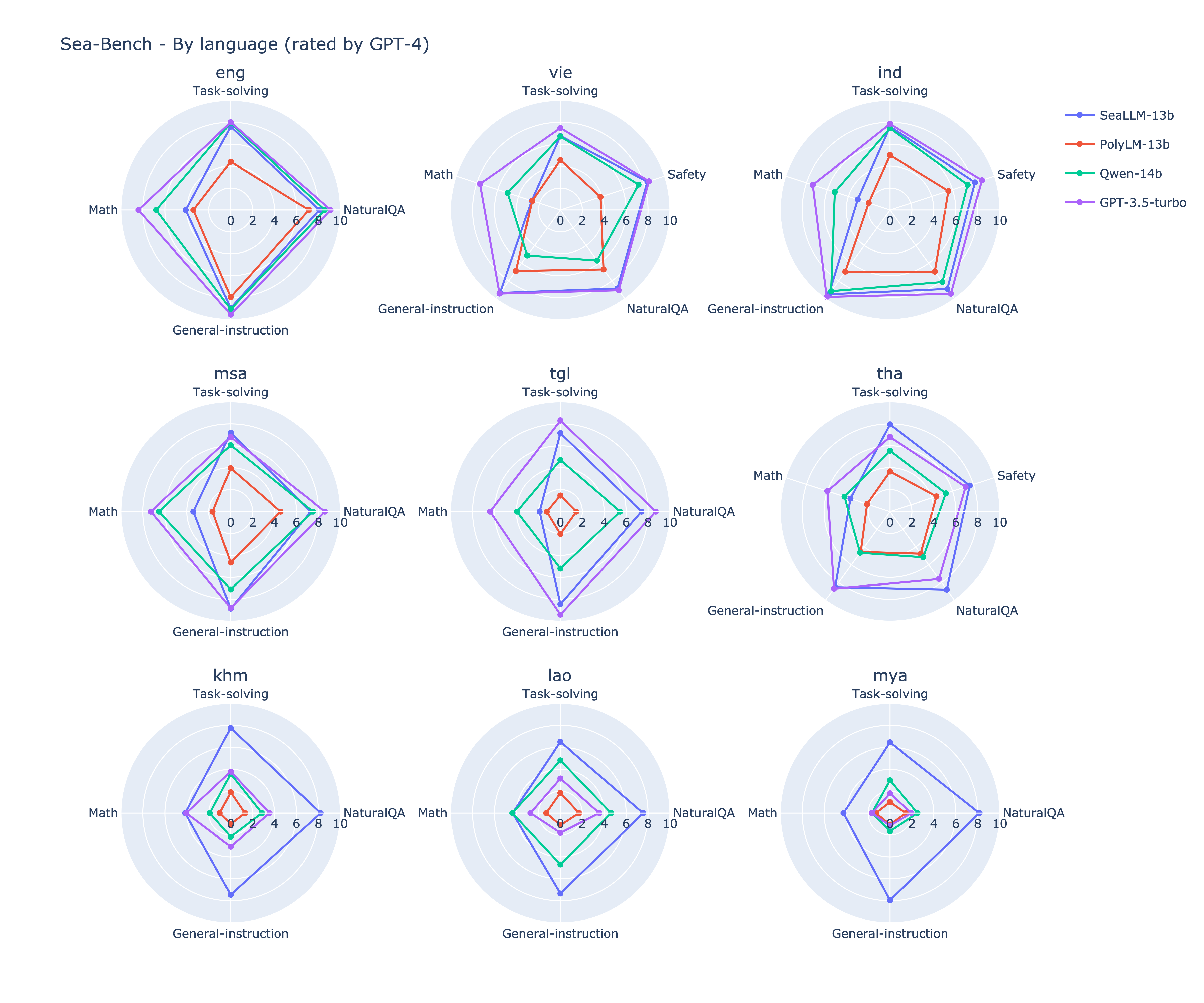} 
    \caption{Sea-bench scores as evaluated by GPT-4 for different models across 9 languages and 5 categories.}
    \label{fig:sea_bench:seallm_vs_chatgpt_breakdown}
\end{figure*}

\Cref{fig:sea_bench:seallm_vs_chatgpt_breakdown} breaks down the GPT-4 rated Sea-bench score-based evaluations of SeaLLM-13b and other baselines by both language and task category. As shown, our SeaLLM-13b model far exceeds ChatGPT-3.5 in most non-Latin languages, such as Burmese (Mya), Lao and Khmer, though it trails behind this formidable competitor in Latin-based languages, mostly in math reasoning skills.

\end{document}